%% file: main.tex
\documentclass{wscpaperproc}
\usepackage{latexsym}
\usepackage{caption,subcaption}
\usepackage{graphicx}
\usepackage{mathptmx}
\usepackage[T1]{fontenc}
\usepackage{multirow}
\usepackage{xcolor}

\usepackage{amsmath}
\usepackage{amsfonts}
\usepackage{amssymb}
\usepackage{amsbsy}
\usepackage{amsthm}

\usepackage[pdftex,colorlinks=true,urlcolor=blue,citecolor=black,anchorcolor=black,linkcolor=black]{hyperref}

\usepackage[noend]{algpseudocode}
\usepackage{algorithm, booktabs}
\newsavebox{\arrangebox}

\algrenewcommand\algorithmicrequire{\textbf{Input:}}
\algrenewcommand\algorithmicensure{\textbf{Output:}}

\newtheoremstyle{wsc}
{3pt}
{3pt}
{}
{}
{\bf}
{}
{.5em}
{}

\theoremstyle{wsc}

\begin{document}

%
%

\pagestyle{fancyplain}

\thispagestyle{plain}
\firstPageHead{}

\chead{\fancyplain{}{\itshape Amaranath, Bhide, Jensen, and Haas}}

\rhead{}
\cfoot{}
\renewcommand{\headrulewidth}{0pt} 

\input{wscbib.tex}           

\setlength{\baselineskip}{12.7pt}

\title{EXTENDING CAUSAL METAMODELING TO A NON-MARKOVIAN QUEUE}

\author{\begin{center}Pracheta Amaranath\textsuperscript{1}, Anant Bhide\textsuperscript{1}, David Jensen\textsuperscript{1} and Peter J. Haas\textsuperscript{1}\\
[11pt]
\textsuperscript{1} Manning College of Information and Computer Sciences,\\ University of Massachusetts Amherst, MA, USA\\
\end{center}
}

\maketitle

\vspace{-12pt}

\section*{ABSTRACT}
Metamodels for discrete-event simulations approximate the behavior of simulation models without running expensive simulations. Prior work introduced modular dynamic Bayesian networks (MDBNs)---a class of metamodels that can estimate a range of probabilistic and causal queries (PCQs) using a single, trained model---but the method was limited to Markovian systems. In this paper, we initiate an extension of MDBNs to non-Markovian queues by approximating non-exponential distributions using phase-type distributions. This approach raises novel challenges, including balancing metamodeling accuracy and tractability when choosing the number of phases, efficiently learning metamodel parameters, and choosing the sampling interval that is used to approximate a continuous-time simulation by a discrete-time MDBN. We provide preliminary solutions to these challenges, yielding the first causal metamodeling technique for non-Markovian systems. Experiments on a G/M/1 queue demonstrate that the MDBN can produce accurate answers to PCQs with orders-of-magnitude speedup of inference times relative to direct simulation.

\section{INTRODUCTION}
\label{sec:introduction}
Simulation metamodeling, in which a complex simulation is approximated via a statistical model that captures the simulation's input-output relationships, is widely used in applications where prediction and optimization must be done quickly, such as in ``system'' and ``process'' digital twin settings; see~\citeN{AWS25}. Modular dynamic Bayesian networks (MDBNs) were recently introduced as the first class of causal simulation metamodels capable of answering a rich set of probabilistic and causal queries (PCQs) from a single trained metamodel~\shortcite{amaranath2023causal}. A PCQ requests computation of a specified conditional probability distribution of the system state at one or more time points, given observations of, and \emph{interventions on}, other state variables and input parameters. For example, in a queueing model, a PCQ might ask: "What would be the distribution of the queue length at 3pm if we were to double the service rate starting at noon?" 

Unlike traditional metamodels which approximate a single real-valued performance measure as a function of input parameters and require a separate model for each query of interest~\cite{barton2020tutorial}, an MDBN represents the joint evolution of simulation state variables over time as a causal graphical model. Based on a set of discrete-time snapshots of the system state, machine learning algorithms can estimate the conditional probability distributions (CPDs) that govern transitions between successive snapshots. Using these CPDs, an MDBN can efficiently estimate answers to many such queries from a single model.
Previously, we demonstrated these capabilities on an M/M/1 queue, showing that a \emph{single} MDBN could accurately answer a variety of PCQs, including queries about interventions not present in the training data.

However, the M/M/1 queue---and Markovian queueing systems more broadly---represent a very restricted class of simulation models. The Markovian property of the M/M/1 queue maps directly onto the two-time-slice structure of an MDBN: the conditional distribution of the state at the next time step is fully determined by the state at the current time step. Many realistic simulation models, on the other hand, employ general service or inter-arrival time distributions (e.g., Gamma, Weibull, or lognormal) that violate this property. Such distributions arise naturally in practice: call centers often exhibit lognormally distributed service times~\shortcite{brown2005statistical}, healthcare systems involve procedure durations that follow Gamma or Weibull distributions~\shortcite{sahu1997weibull} and project management systems are typically modeled using Beta distributions~\cite{keefer1993better}. In these settings, the system carries memory---for example, in a G/G/1 queue with Gamma-distributed service times, the time remaining until the next service completion depends not only on whether the server is busy, but on how long the current service has been in progress. This duration dependence means that the queue-length process at the next time step cannot be determined from the current queue length alone, breaking the structural assumption underlying the original MDBN. Extending MDBNs to such non-Markovian systems is essential for the approach to be applicable to the broad class of simulation models encountered in practice.

In this paper, we initiate the extension of MDBNs to non-Markovian queueing simulations in the setting of a G/M/1 queue.
Our approach first approximates the general (non-Markovian) inter-arrival distribution with a phase-type distribution---distributions constructed from sequences and mixtures of exponential stages that satisfy the Markov property by design. 
We then augment the structure of the MDBN with phase variables that track the state of the arrival process, recovering the two-time-slice dependency structure pertaining to the Markov assumption. This ``method of phases'' approach has long been used to allow numerical computation of probabilities for non-Markovian systems---see, e.g., \cite{neuts1994matrix}---but has not been apllied in the metamodeling setting.

This augmentation, however, increases both the dimensionality of the state space and the number of conditional probability distribution (CPD) parameters to be estimated, which creates a scalability challenge. We address this challenge by: (i)~adopting phase-distribution approximation methods that balance the accuracy of the approximation with computational efficiency; (ii)~introducing a technique called ``parameter extrapolation''---a novel specialization of the general ``parameter tying'' method from the graphical models literature~\cite[Section 17.5]{koller2009probabilistic}---which exploits structural regularities in the queue-length CPDs to reduce the number of parameters estimated from data; and (iii)~choosing the MDBN ``sampling interval'' to balance accuracy and computational efficiency. This latter issue arises because a discrete-event simulation generates continuous-time sample paths but an MDBN operates in discrete time, i.e., over a sequence of system-state snapshots. In the M/M/1 setting of \shortciteN{amaranath2023causal}, the sampling interval was selected empirically by evaluating query accuracy across a small set of candidate values. This approach becomes impractical for the G/M/1 queue, where the expanded parameter space---now including the parameters of the phase variables in addition to the service rate---makes exhaustive tuning over all parameter combinations prohibitively expensive. We instead derive the optimal sampling interval analytically, using results from the theory of continuous-time Markov chain discretization~\cite{doytchinov2010time}, which provides a closed-form bound that applies uniformly across all parameter settings. 

We evaluate the MDBN on a workload of PCQs, comparing queue-length distributions conditioned on either prior observations or prior interventions to ground-truth conditional distributions from extensive Monte Carlo simulation of the (modified) G/M/1 queue. Similarly, we compare summary statistics of the conditional distributions---mean queue length and the interquartile range (IQR) of the queue-length distribution---to ground-truth values.


This work makes several novel contributions\footnote{Our code is publicly available at \href{https://github.com/PrachetaBA/non-markovian-mdbn.git}{https://github.com/PrachetaBA/non-markovian-mdbn.git}}. First, we describe structural modifications to the MDBN that accommodate the increased state-complexity of non-Markovian queues while keeping inference tractable. Second, we present methods to improve CPD learning efficiency, including effective approximation by phase-type distributions, parameter extrapolation, and a theoretically grounded approach to selecting the optimal sampling interval. Third, we provide empirical evidence that MDBN metamodels produce accurate answers to probabilistic and causal queries for non-Markovian queues, with inference times that are four orders of magnitude faster relative to direct simulation. To our knowledge, this is the first application of causal graphical model techniques to non-Markovian simulation metamodeling.


\section{BACKGROUND}
\label{sec:background}
\paragraph{Probabilistic and Causal Queries (PCQs)} We briefly review the PCQ framework introduced in \shortciteN{amaranath2023causal}; the reader is referred to that paper for a complete treatment. Consider a discrete-event simulation model $M$ with $l$ state variables whose output is a continuous-time stochastic process $\{X_t\}_{t \geq 0}$ with discrete state space $\mathcal{X} \subseteq \mathbb{Z}^l$, and input parameter vector $\theta = (\theta_1, \ldots, \theta_d) \in \Theta \subseteq \mathbb{R}^d$. A probabilistic and causal query (PCQ) is a desired conditional probability distribution of projections of the system state at one or more time points, conditioned on a set of \emph{actions} applied at specified times or over time intervals. Actions can be \emph{observational}---denoted $Y_t = y$, conditioning on the state taking a particular value---or \emph{interventional}---denoted $Y_t \leftarrow f(Y_t)$, externally setting the state via the $\text{do}(\cdot)$ operator~\cite{pearl2009causality}. For example, in a queueing system with arrival rate $\lambda$, service rate $\mu$, and queue length $L_t$, representative PCQs include: inferring the queue-length distribution under fixed rates ($P(L_\tau \mid \lambda_{0:\tau} \leftarrow c_1, \mu_{0:\tau} \leftarrow c_2)$), reasoning about queue-length distributions after injecting additional customers ($P(L_{\tau_1}, L_{\tau_2} \mid L_{\tau_1-h} \leftarrow L_{\tau_1-h} + 5, \ldots)$), and identifying the arrival rate that maximizes the probability of a target queue-length range ($\arg\max_\lambda P(L_\tau \in [5..10] \mid \lambda)$).

PCQs fall into several subclasses relevant to our experiments. \emph{Interventional queries} involve at least one interventional action on system states or parameters, where the intervention action could force the system state to a fixed value or increment/decrement it based on its current value. \emph{Conditional queries} involve observations on system states or parameters and \emph{inverse queries} optimize over simulation input parameters to achieve a target system behavior. A single trained MDBN can answer all of these query types without requiring separate models or additional simulation runs. Note that interventional queries are usually hard to support via execution of a simulation model without altering the simulation code.

\paragraph{Modular Dynamic Bayesian Networks (MDBNs)} An MDBN is a dynamic Bayesian network (DBN) whose edges represent direct causal relationships and whose conditional probability distributions (CPDs) satisfy the modularity property: interventions on any variable do not affect the CPDs of other variables. These properties allow the MDBN to answer interventional queries by translating them into equivalent probabilistic queries via standard techniques such as the backdoor criterion; see~\shortciteN{amaranath2023causal} for a detailed treatment of the causal assumptions, inference procedures, and the algorithms used to implement interventions.

\begin{figure}[htbp]
     \begin{subfigure}[b]{0.2\textwidth}
         \centering
         \includegraphics[scale=0.4]{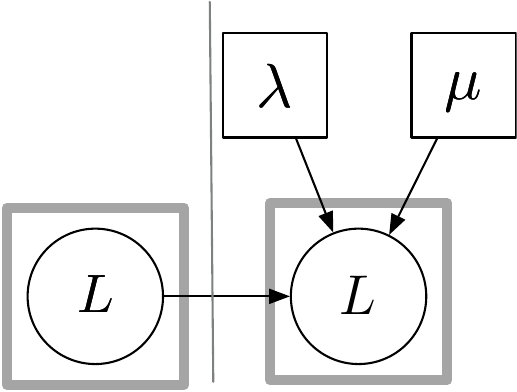}
         \caption{MDBN plate model}
         \label{fig:mdbn-mm1-plate}
    \end{subfigure}
    \hspace{0.5cm}
    \begin{subfigure}[b]{0.28\textwidth}
        \small
        \centering
        \begin{tabular}{ccccc}
        \toprule
           $\lambda$ & $\mu$ & $L^{(j)}$ & $L^{(j+1)}$ & $P(L^{(j+1)} \mid \lambda, \mu, L^{(j)})$  \\ \midrule
           $0.1$ & $0.4$ & $0$ & $[0..10]$ & $[0.9...0.00]$ \\
           $0.1$ & $0.9$ & $0$ & $[0..10]$ & $[0.7...0.01]$ \\
           $\vdots$ & $\vdots$ & $\vdots$ & $\vdots$ & $\vdots$ \\
           $0.3$ & $0.9$ & $10$ & $[0..10]$ & $[0.0...0.3]$ \\
           \bottomrule
        \end{tabular}
        \caption{Example CPD}
        \label{tab:mm1-cpd}
    \end{subfigure}
    \hspace{2.7cm}
    \begin{subfigure}[b]{0.26\textwidth}
         \centering
         \includegraphics[scale=0.4]{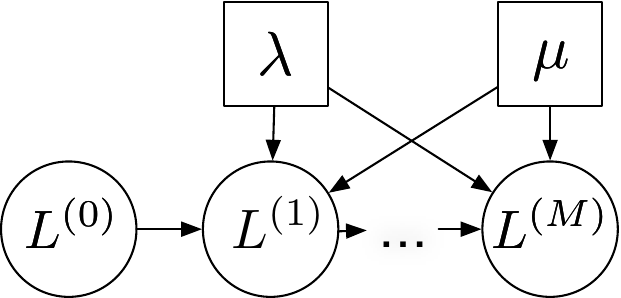}
         \caption{MDBN ground network $\mathbb{G}$}
         \label{fig:mdbn-mm1-ground}
     \end{subfigure}
    \caption{MDBN model for the M/M/1 queue}
    \label{fig:mdbn-mm1-structure}
\end{figure}

For the M/M/1 queue studied in prior work, the MDBN consists of three variables: arrival rate $\lambda$, service rate $\mu$, and queue length $L$. The structure of the MDBN is depicted in Figure~\ref{fig:mdbn-mm1-plate}---the input parameters $\lambda$ and $\mu$ are parents of $L$ in every time slice, i.e., each discrete time step at which the system state is recorded. The queue length at each time slice ($j+1$) depends on the queue length at the previous time slice ($j$) and the input parameters: $P(L^{(j+1)} \mid \lambda, \mu, L^{(j)})$. The thick gray box around $L$ indicates that this dependency is repeated across time slices; unrolling the plate model for $M$ slices yields the ground network $\mathbb{G}$ shown in Figure~\ref{fig:mdbn-mm1-ground}. This structure is a direct consequence of the Markov property of the M/M/1 queue-length process---the distribution of the future queue length is fully determined by the current queue length and the input parameters, with no dependence on the history. The CPD for $P(L^{(j+1)} \mid \lambda, \mu, L^{(j)})$ (see Figure~\ref{tab:mm1-cpd}) is learned from simulation data by sampling the queue length at each time point $\delta_j$ and computing maximum likelihood estimates, where $\delta_j = j \cdot \delta$ and $\delta$ is the sampling interval controlling the temporal resolution at which the MDBN is constructed.

Two design parameters influence the quality of the learned MDBN---the sampling interval $\delta$ and the number of simulation runs $N$ used to learn the CPD parameters. Prior work showed empirically that when $\delta$ is a small fraction of the time scale $\delta_Q$ of the query of interest, the discrete-time MDBN accurately captures the continuous-time dynamics; in this paper, we provide a theoretical basis for selecting the optimal $\delta$ (Section~\ref{sec:optimal-delta}). The number of simulation runs $N$ determines the accuracy of the estimated CPDs, with diminishing returns beyond a sufficient threshold. Both parameters involve tradeoffs between accuracy and computational cost that carry over to the non-Markovian extension for MDBNs.

\paragraph{Phase-Type Distributions} A phase-type distribution describes the time to absorption in a finite-state continuous-time Markov chain (CTMC) with a single absorbing state. Intuitively, a random variable with a phase-type distribution is generated by passing through a sequence of exponential stages (or \emph{phases}), where transitions between stages occur at exponential rates. Because the holding time in each phase is memoryless, the overall process is Markovian by construction---even though the resulting distribution of the total time to absorption can approximate a wide variety of non-exponential shapes. This property is what makes phase-type distributions useful for our purposes: by replacing a general (non-Markovian) distribution with a phase-type approximation, we recover the Markov structure required by the MDBN. The class of phase-type distributions is dense in the set of all nonnegative distributions~\cite{neuts1994matrix},
meaning that any such distribution can be approximated arbitrarily well given a sufficient number of phases. In practice, however, the number of phases directly impacts the complexity of the resulting MDBN, so we seek approximations that use as few phases as possible while maintaining adequate fidelity.

The phase-type distribution that we employ is the \emph{generalized Erlang distribution} (GED), also known as a hypoexponential distribution, which consists of $k$ phases in series with distinct rates $\lambda_1, \lambda_2, \ldots, \lambda_k$. The total time to absorption is the sum of $k$ independent exponential random variables, each with its own rate. When all rates are identical, this reduces to an Erlang distribution. The specific methods used to obtain GED approximations for the Gamma, Beta, and Weibull distributions are described in Section~\ref{sec:phase-type-approximation}. 

\section{MODULAR DYNAMIC BAYESIAN NETWORKS FOR NON-MARKOVIAN QUEUES}
\label{sec:mdbn-non-markovian}
We describe how the MDBN framework can be extended to the G/M/1 queue. Specifically, we note the challenges posed by non-Markovian inter-arrival distributions, the phase-type approximations used to recover a Markovian representation, the resulting MDBN structure, an efficient approach to CPD estimation, and a theoretically grounded method for selecting the optimal sampling interval. 

\subsection{The G/M/1 Queue}
\label{sec:gm1-queue}
We consider a G/M/1 queue in which the inter-arrival times follow a general distribution $F_A$ and the service times are exponentially distributed with rate $\mu$. The generality of $F_A$ is what distinguishes this setting from the M/M/1 queue studied in prior work: $F_A$ may be any distribution on $[0,\infty)$ that does not necessarily satisfy the Markov property, such as the Gamma, Beta, or Weibull. We select specific parameterizations of these distributions in our experiments to cover both light-tailed and heavy-tailed regimes. While we focus on the G/M/1 queue as a proof of concept, the techniques described in this section apply to G/G/1 queues more broadly.

The key challenge in determining the MDBN structure is that, unlike the M/M/1 queue, the queue length at time $t + \delta$ depends not only on the queue length at time $t$ and simulation input parameters, but also on the elapsed time since the last arrival. This dependence has two consequences for the MDBN: the queue-length process is no longer time-homogeneous in the original state space, and the two-time-slice structure of the original MDBN is insufficient to capture these dynamics, since it assumes that the conditional distribution $P(L^{(j+1)} \mid L^{(j)}, \lambda, \mu)$ is the same for every time slice. To address both issues, we augment the state space with variables that track the phase of the inter-arrival process, effectively encoding the relevant dependency in a Markovian form. This augmentation is made possible by the phase-type approximation described next.

\subsection{Phase-Type Approximation of General Distributions}
\label{sec:phase-type-approximation}
To extend the MDBN framework to a non-Markovian queue, we approximate the general inter-arrival distribution by a phase-type distribution that can be incorporated into the MDBN. In this paper, we approximate the Gamma, Weibull, and Beta distributions using GEDs.

\begin{figure}[htbp]
    \centering
    \begin{subfigure}[b]{0.32\textwidth}
        \centering
        \includegraphics[width=\textwidth]{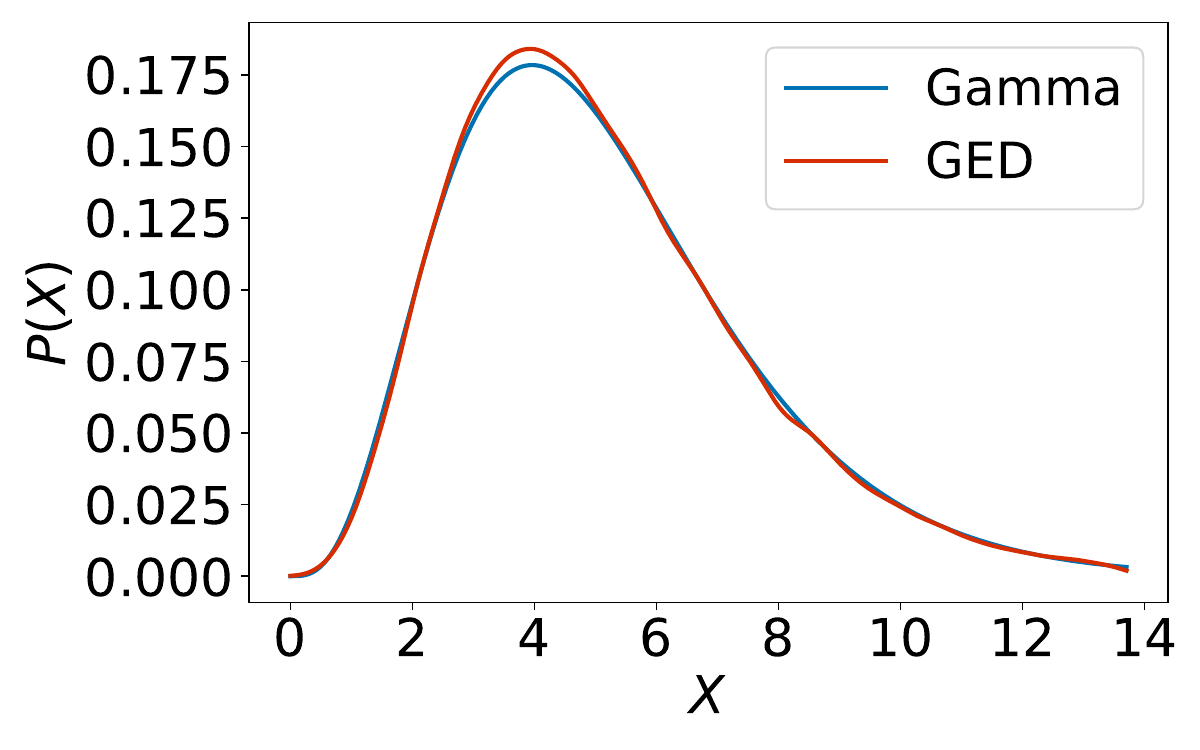}
        \caption{\centering Gamma: $\alpha = 4.3, \beta = 1.2$;\\ GED: $k = 5, \lambda_{1..3} = 0.97, \lambda_4 = 0.59, \lambda_5 = 2.68$; JSD: $0.0022$}
        \label{fig:phase-type-gamma}
    \end{subfigure}
    \hfill
    \begin{subfigure}[b]{0.32\textwidth}
        \centering
        \includegraphics[width=\textwidth]{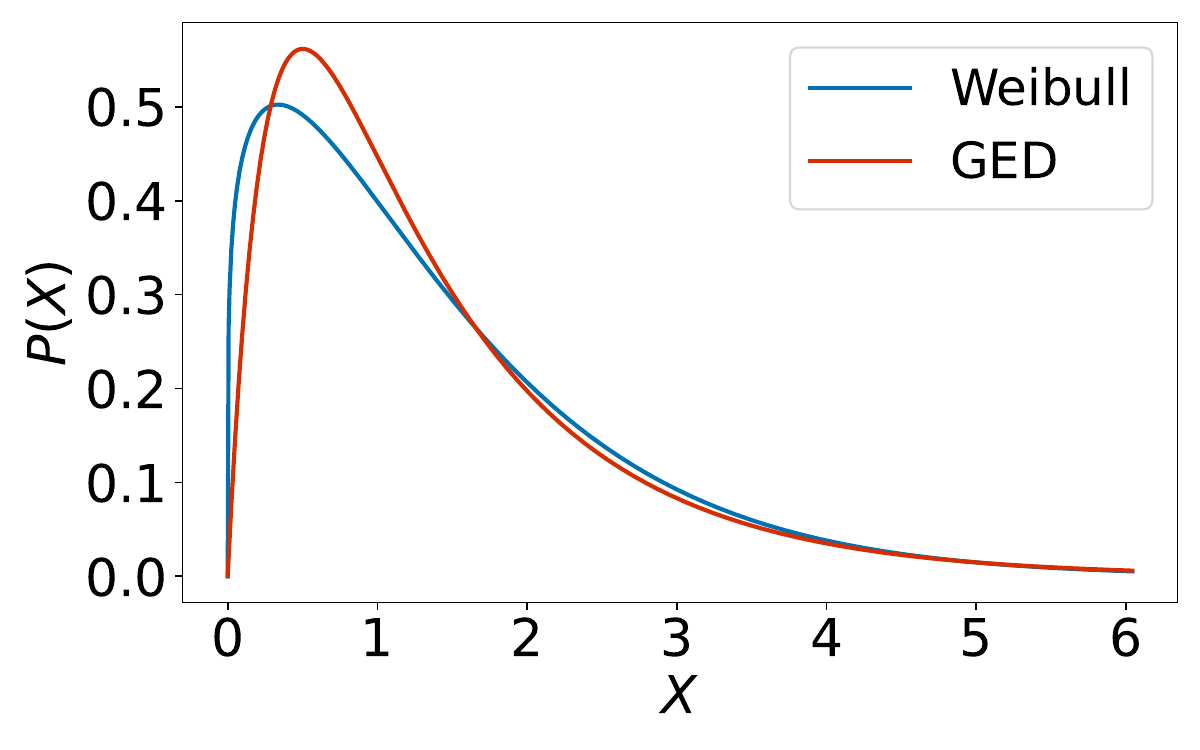}
        \caption{\centering Weibull: $\alpha=1.2, \beta=1.5$\\ GED: $k = 2, \lambda_1 = 0.87, \lambda_2 = 3.8$; JSD: $0.0040$}
        \label{fig:phase-type-weibull}
    \end{subfigure}
    \hfill
    \begin{subfigure}[b]{0.32\textwidth}
        \centering
        \includegraphics[width=\textwidth]{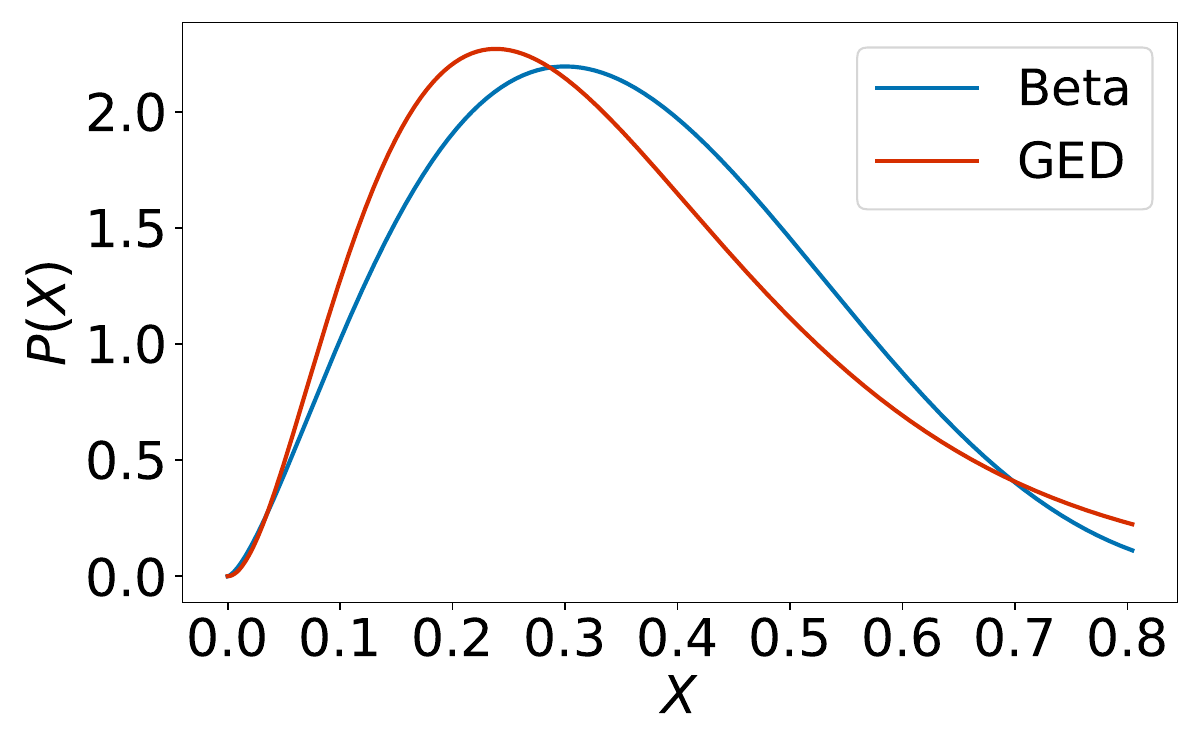}
        \caption{\small\centering Beta: $\alpha = 2.5, \beta = 4.5$\\ GED: $k = 3, \lambda_1 = 8.39, \lambda_2 = 8.39, \lambda_3 = 8.39$; JSD: $0.0043$}
        \label{fig:phase-type-beta}
    \end{subfigure}
    \caption{Phase-type (GED) approximations of target distributions for the G/M/1 queue. Each plot shows the target distribution (blue) and the GED approximation (red). $\alpha$ and $\beta$ denote the corresponding (shape/scale) parameters of the target distribution. The Jensen-Shannon distance (JSD) indicates the quality of the fit.}
    \label{fig:phase-type-fits}
\end{figure}

\paragraph{Gamma Distribution} For the Gamma distribution, we use the closed-form GED construction of~\shortciteN{cassidy2022numerical}. Given $X \sim \Gamma(\alpha, \beta)$, the approximation concatenates $k$ independent exponential phases: $k - 2$ phases share a common rate, while the remaining two have distinct rates. The phase rates are determined using a closed-form, analytical solution that matches the first two moments of the target distribution. $k$ is determined as $\lceil \alpha \rceil$ for $\alpha > 2$. In our experiments with the Gamma distribution, we restrict $\alpha < 5$.  
An example fit is shown in Figure~\ref{fig:phase-type-gamma}. 

\paragraph{Weibull and Beta Distributions}  For the Weibull and Beta distributions, closed-form GED approximations are not available. We use the moment-matching approach of~\shortciteN{elmagrhraby2009approximation}, which equates the first $n$ moments of the target distribution with those of a GED with $n$ phases, yielding a system of $n$ nonlinear equations in the phase rates $\lambda_1\ldots\lambda_n$. When this system is infeasible---as often occurs for three or more phases---the method formulates a mathematical program that minimizes weighted deviations from the target moments. We sweep over candidate values of $k \in [2..10]$ and select the best fit using the Jensen-Shannon divergence, with cross-entropy as a tie-breaker. Representative fits are shown in Figures~\ref{fig:phase-type-weibull} and~\ref{fig:phase-type-beta}. As shown in Section~\ref{sec:experiments}, restricting the maximum candidate values of $k \leq 10$ is sufficient to achieve low error for our MDBN metamodels. 

\subsection{MDBN Structure}
\label{sec:mdbn-structure}
To represent the state of a G/M/1 queue with a phase-type inter-arrival distribution, we augment the MDBN with a random variable $S$ denoting the current phase of the arrival process. The queue length at the time $t + \delta$ depends on the state of the arrival process within its sequence of exponential phases, since this determines the probability of an arrival occurring before the next time slice.

\begin{figure}[htbp]
     \centering
     \begin{subfigure}[b]{0.2\textwidth}
         \centering
         \includegraphics[scale=0.18]{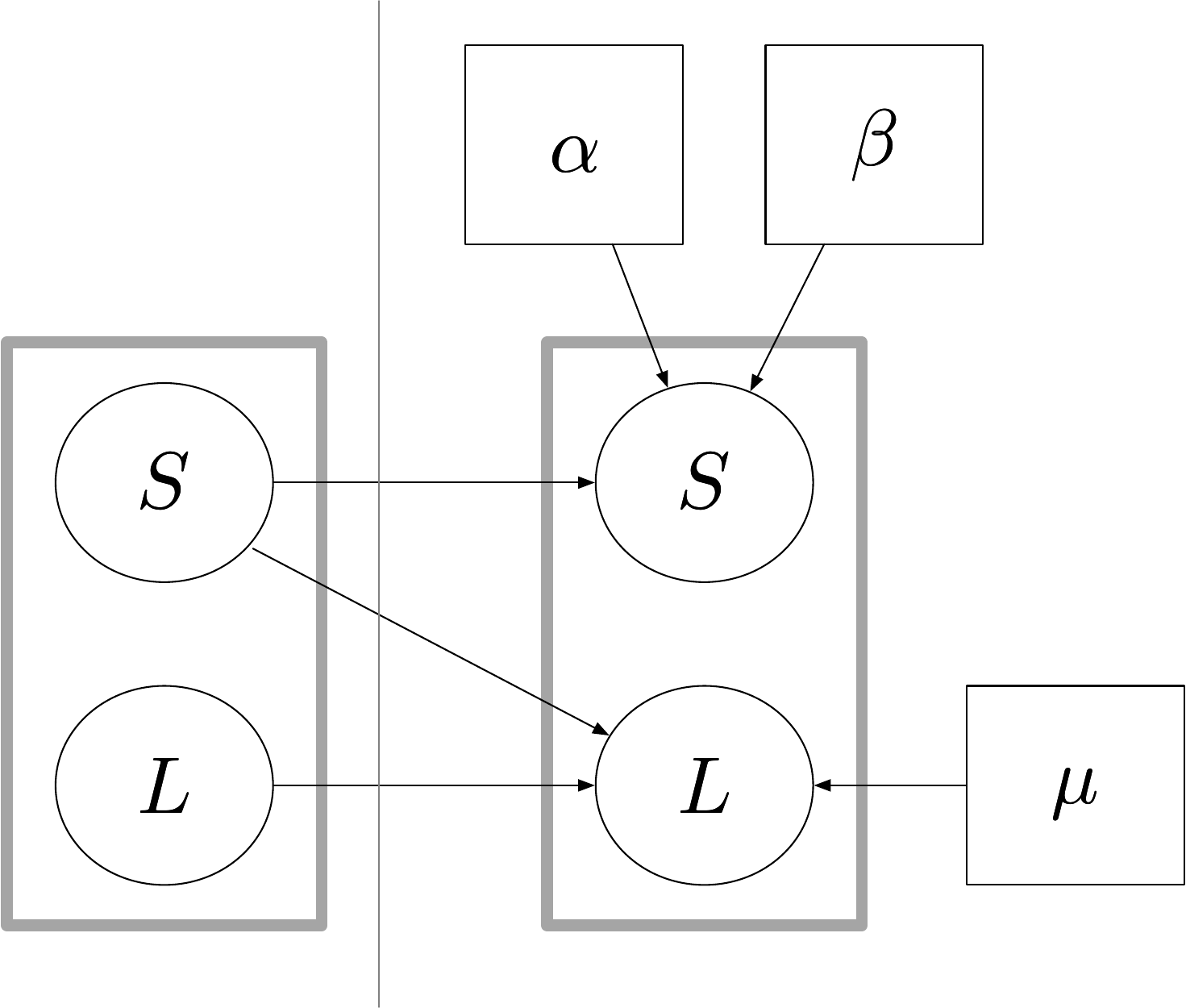}
         \caption{MDBN plate model}
         \label{fig:mdbn-plate}
    \end{subfigure}
    \hspace{1.5cm}
    \vline
    \hspace{0.1cm}
    \begin{subfigure}[b]{0.2\textwidth}
         \centering
         \includegraphics[scale=0.17]{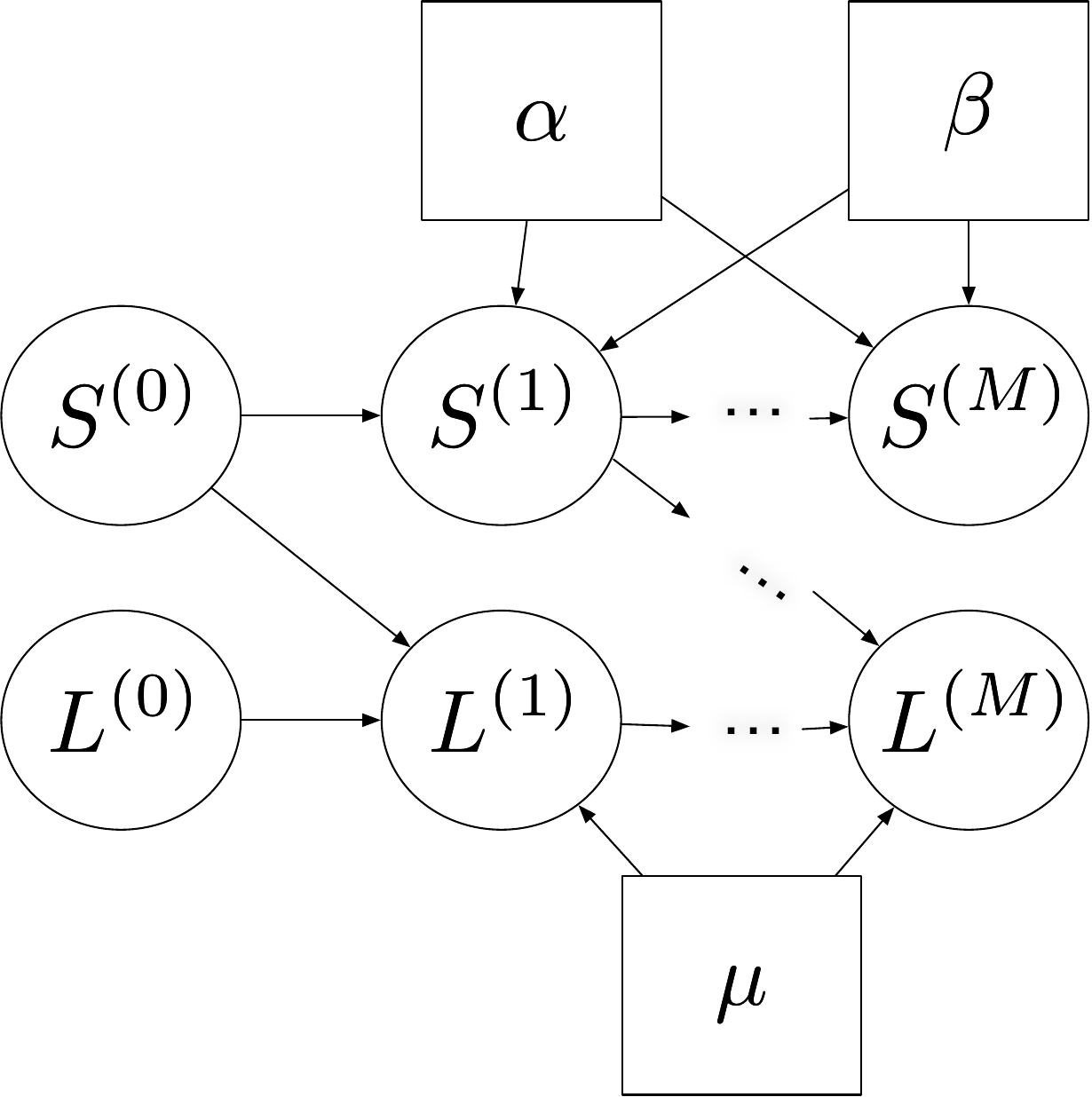}
         \caption{MDBN ground network $\mathbb{G}$}
         \label{fig:mdbn-ground}
     \end{subfigure}
    \caption{MDBN model for the G/M/1 queue}
    \label{fig:mdbn-structure}
\end{figure}

The modified MDBN therefore contains two dynamic variables inside the plate: the queue length $L$ and the phase variable $S$. The phase $S^{(j+1)}$ depends on the phase $S^{(j)}$ and the parameters of the general inter-arrival distribution, denoted $\omega = (\alpha, \beta)$ for the Gamma, Beta, and Weibull distributions. The queue length $L^{(j+1)}$ depends on the queue length $L^{(j)}$, the phase $S^{(j)}$, as well as the service rate $\mu$. The phase variable $S$ implicitly captures the dependence on the GED phase rates $\lambda_1, \ldots, \lambda_k$, which are deterministic functions of $\omega$. Therefore, we do not explicitly model them as random variables in the MDBN
and just take them into account when computing CPDs.
The plate model and the corresponding (unrolled) ground network $\mathbb{G}$ for $M$ time slices are shown in Figure~\ref{fig:mdbn-structure}.

\subsection{Optimal Sampling Interval}
\label{sec:optimal-delta}
In prior work that modeled the M/M/1 queue with MDBNs, the sampling interval $\delta$ was selected empirically by evaluating PCQ accuracy across a range of input parameter values. This was feasible because the M/M/1 queue has only two input parameters $(\lambda, \mu)$, making the computation tractable. As we move toward G/M/1 (and potentially G/G/1) queues, the number of input parameters grows---each phase-type approximation introduces additional phase rates---and empirical tuning over all parameter combinations becomes impractical. We therefore provide a principled method for computing $\delta$ based on the theory of discretizing continuous-time Markov chains (CTMCs).

Every queueing process has an underlying CTMC whose states correspond to the possible configurations of the queue. For non-Markovian queues, this requires augmenting the state space to include the phase of the arrival or service process, as described in Section~\ref{sec:mdbn-structure}. Observing this CTMC at equally spaced time points $\delta_j = j \cdot \delta$ yields a discrete-time Markov chain (DTMC) with one-step transition matrix $P(\delta)$. In the M/M/1 case, each entry represents the probability $P(L^{(j+1)} = \ell' \mid L^{(j)} = \ell)$ of transitioning from queue length $\ell$ at time $\delta_j$ to queue length $\ell'$ at time $\delta_{j+1} = \delta_j + \delta$. The key observation is that the CPDs of the MDBN encode these same transition probabilities, conditioned additionally on the input parameters $\theta$. The two representations are therefore equivalent for any fixed $\theta$, and any discretization guarantee that holds for the DTMC applies directly to the MDBN.

Prior work on optimal discretization of CTMCs~\cite{doytchinov2010time} shows that for a second-order approximation of the matrix exponential $P(t) = e^{Qt}$, the sampling interval satisfying
\[ 0 < \delta \leq \frac{1}{\max_i |q_{ii}|} \]
ensures that $P(\delta)$ is a valid transition matrix with bounded error. Here, $q_{ii}$ are the diagonal elements of the infinitesimal generator matrix $Q$ corresponding to the CTMC. Since the MDBN must produce accurate CPDs across all combinations of input parameters, the optimal $\delta$ is obtained by minimizing over the full parameter space:
\begin{equation}\label{eq:optimal-delta}
    \delta_{\text{optimal}} = \min_{\theta \in \boldsymbol{\Theta}} \left( \frac{1}{\max_i |q_{ii}(\theta)|} \right)
\end{equation}
where $q_{ii}(\theta)$ are the diagonal elements of $Q$ for parameter values $\theta$ and $\boldsymbol{\Theta}$ is the set of all parameter combinations used in the MDBN. For the M/M/1 queue, $\theta = (\lambda, \mu)$ and $\max_i |q_{ii}| = \lambda + \mu$, so that $\delta_{\text{optimal}} = \frac{1}{\max_{\theta \in \boldsymbol{\Theta}} (\lambda + \mu)}$. This generalizes to the G/M/1 queue, where the Q-matrix is constructed from the GED phase rates $\lambda_1, \ldots, \lambda_k$ and the service rate $\mu$. This also makes explicit a tradeoff in the phase-type approximation: a larger number of phases increases $\max_i |q_{ii}|$, which decreases $\delta_{\text{optimal}}$, leading to more time slices and longer inference times. This is why we seek GED approximations that use as few phases as possible while maintaining adequate fidelity to the target distribution.

\subsection{Estimating the conditional probability distributions (CPDs)}
\label{sec:estimating-cpds}
Having specified the MDBN structure, we estimate two CPDs from simulation data: $P(S^{(j+1)} \mid S^{(j)}, \omega)$ governing the phase transitions of the arrival process, and $P(L^{(j+1)} \mid L^{(j)}, S^{(j)}, \mu)$ governing the queue-length transitions. $P(S^{(j+1)} \mid S^{(j)}, \omega)$ is estimated directly via maximum likelihood estimation (MLE). 
Although $\alpha, \beta$ and $\mu$ are fixed simulation input parameters set by the analyst, we treat them as Bayesian random variables during MDBN training. Their domains are restricted to be the values specified by the experimental design, and their corresponding CPDs $P(\alpha)$, $P(\beta)$, and $P(\mu)$ are set to a discrete uniform distribution over the design points; see \shortciteN{amaranath2023causal}. In our experiments, we use a full-factorial design with $N$ simulation runs per design point. 
The initial phase distribution $P(S^{(0)} = 1) = 1.0$ is deterministic, reflecting that every arrival process begins in its first phase. The initial queue-length distribution $P(L^{(0)})$ is learned from the training data; as in the M/M/1 queue, we vary $L_0 \in \{0..5\}$ across runs to improve estimation of queue length probabilities $P(L \geq 10)$. The simulation time horizon is set to $T = 20.0$ in our experiments.

Training data is generated by running the simulation $N = 1000$ times for each design point, sampling the resulting trajectories at the optimal interval $\delta_{\text{optimal}}$ (Section~\ref{sec:optimal-delta}), and recording $L^{(j)}$ and $S^{(j)}$ at each time slice. Because the phase-type augmentation recovers the Markov property, the augmented process is time-homogeneous, allowing us to pool data across time slices. The number of CPD parameters grows exponentially with the cardinality of the parent variables: the domain of $S$ is determined by the number of GED phases, and the domain of $L$ by the maximum observed queue length across all simulation runs. In our experiments, $|\text{dom}(\alpha)| = |\text{dom}(\beta)| = |\text{dom}(\mu)| = 2$.

\begin{figure}[htbp]
    \centering
    \includegraphics[width=\textwidth]{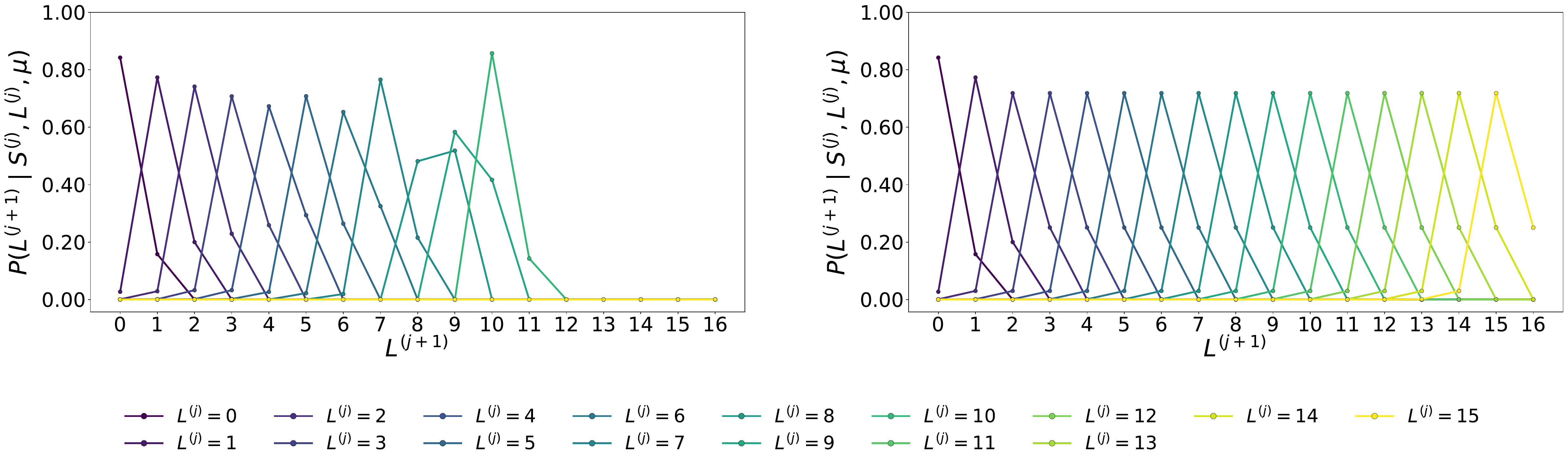} 
    \caption{(left) CPD parameters estimated using MLE---note the lack of training data for $L > 13$; (right) CPD parameters estimated using parameter extrapolation. For $\ell > 1$, the conditional distribution $P(L^{(j+1)} \mid L^{(j)} = \ell, S^{(j)}, \mu)$ is approximately a translation of the distribution for $\ell - 1$.}
    \label{fig:param-extrapolation}
\end{figure}

\paragraph{Parameter Extrapolation} The CPD $P(L^{(j+1)} \mid L^{(j)}, S^{(j)}, \mu)$ must be estimated for every combination of parent values. For the boundary cases $L^{(j)} = 0$ and $L^{(j)} = 1$, the CPD entries are estimated directly via MLE. However, for $\ell > 1$, we observe that for fixed parent variable values, excluding $L^{(j)}$, the conditional distribution $P(L^{(j+1)} \mid L^{(j)} = \ell, S^{(j)}, \mu)$ exhibits a characteristic pattern: the shape of the distribution remains the same across different values of $\ell$, but is shifted to the right by an offset corresponding to $\ell$. That is, the distribution for $L^{(j)} = \ell$ is approximately a translation of the distribution for $L^{(j)} = \ell - 1$, shifted by one unit. This pattern is illustrated in Figure~\ref{fig:param-extrapolation}.

We exploit this regularity through a technique we call \emph{parameter extrapolation}. Rather than estimating a separate distribution for each value of $\ell > 1$, we learn a single distribution from the pooled data across all $\ell > 1$ and then populate the CPD parameters by applying the appropriate shift for each $\ell$. This provides two advantages: it pools data across queue-length values, improving the accuracy of the estimates, and it reduces the number of parameters that need to be estimated independently. The procedure is formalized in Algorithm~\ref{alg:parameter-extrapolation}, where $\theta = (s, m)$ are specified values of parents $(S^{(j)}, \mu)$ of $L^{(j+1)}$, $\boldsymbol{\Theta}$ is the set of all $\theta$'s considered, and $D_{\theta}$ is the training data conditioned on $\theta$ and sampled at interval $\delta_{\text{optimal}}$; here $D_\theta$ comprises $N$ replications, each with $J$ time slices. Moreover, $L^{(i)}_{r,\theta}$ denotes the queue length observed at the $i$th time slice in the $r$th replication in $D_{\theta}$ and $I(\cdot)$ denotes the indicator function.


\begin{algorithm}[htb]
\caption{Parameter Extrapolation}\label{alg:parameter-extrapolation}
\begin{algorithmic}[1]
\State \textbf{Input:} $\boldsymbol{\Theta}$, $\{D_{\theta}: \theta \in \boldsymbol{\Theta}\}$, $\ell_\text{max}$
\State \textbf{Output:} Extrapolated CPDs $\hat{P}(L^{(j+1)} \mid L^{(j)}, \theta)$, $\theta\in\boldsymbol{\Theta}$
\For{each parent assignment $\theta \in \boldsymbol{\Theta}$}\quad\textcolor{gray}{// Compute CPDs for $\ell\in\{0,1\}$:}
\State $Q(\ell, \ell') \leftarrow (NJ)^{-1} \sum_{r=1}^{N} \sum_{j=0}^{J-2} I(L_{r,\theta}^{(j)} = \ell, L_{r,\theta}^{(j+1)} = \ell')$,\hfill $\ell \in \{0, 1\}$, $\ell' \in [0..\ell_{\text{max}}]$
    \State $Q(\ell' \mid \ell) \leftarrow Q(\ell, \ell') / \sum_{\ell'=0}^{\ell_{\text{max}}} Q(\ell, \ell')$,\hfill$\ell \in \{0, 1\}$, $\ell' \in [0..\ell_{\text{max}}]$
    \State \textcolor{gray}{// Pool data over $\ell\in[2..\ell_\text{max}]$ (here $\Delta$ is the queue-length change):}
    \State $R(\Delta) \leftarrow \bigl(NJ(\ell_\text{max}-1)\bigr)^{-1} \sum_{r=1}^{N} \sum_{j=0}^{J-2} \sum_{\ell = 2}^{\ell_{\text{max}}} I(L_{r,\theta}^{(j)} = \ell, L_{r,\theta}^{(j+1)} = \ell+\Delta)$,\hfill $\Delta \in [-2..(\ell_{\text{max}}-2)]$
    \State \textcolor{gray}{// Compute CPDs:}
    \For{$\ell,\ell' \in [0..\ell_{\text{max}}]$}
    \State $\hat{P}(L^{(j+1)} = \ell' \mid L^{(j)} = \ell, \theta) = 
    \begin{cases} 
    Q(\ell' \mid \ell) & \text{if $\ell \in \{0, 1\}$ and $\ell' \in [0..\ell_{\text{max}}]$} \\
    R(\ell'-\ell) & \text{if $\ell \in [2..\ell_{\text{max}}]$ and $\ell'-\ell\in [-2..\ell_{\text{max}}-2]$} \\
    0 & \text{otherwise}
    \end{cases}$
\EndFor
\EndFor
\State \Return $\hat{P}$
\end{algorithmic}
\end{algorithm}

\subsection{Inference}
\label{sec:exact-inference}

Given the learned MDBN, we compute answers to PCQs as in the M/M/1 queue: interventional queries are first converted to equivalent probabilistic queries via the TRANSFORM algorithm described in~\shortciteN{amaranath2023causal}, and the resulting queries are answered using standard inference on the ground network. In our experiments, we use lazy propagation~\cite{madsen1999lazy}, an exact inference algorithm for Bayesian networks, implemented in the \texttt{pyAgrum} library~\shortcite{ducamp2020pyagrum}. Despite the increased size of the network due to phase variable $S$, inference times remain on the order of milliseconds per query.

\section{EXPERIMENTAL RESULTS}
\label{sec:experiments}

We evaluate the MDBN metamodel of a G/M/1 queue with three general inter-arrival distributions: Gamma, Beta, and Weibull. 
For each distribution, we assess the accuracy of the MDBN's answers to a variety of PCQs by comparing them to the estimated ground-truth probability distributions. We first describe the simulation input parameters that comprise the query workload, the evaluation metrics, and finally present the results for each distribution. 

\subsection{Query Workload}
\label{sec:query-workload}

We construct a query workload $W$ comprising four types of PCQs for the G/M/1 queue, totaling 500 queries. For each query, the intervention and query times $\tau_i$ and $\tau_j \leq T$ are set to multiples of $\delta_{\text{optimal}}$, and the values of the simulation input parameters and queue lengths are sampled from their respective domains. $T$ denotes the time horizon for all queries in the workload $W$. We choose input parameters $\{\omega, \mu\}$ such that the queue is stable in $t \in [0..T]$. 
Although each query below involves a single conditional or interventional event, the MDBN framework can handle multiple simultaneous interventions. The four types of PCQs are outlined below.

\paragraph{Conditional queries} These queries infer the queue-length distribution at time $\tau_j$ given an observation of the queue length at an earlier time $\tau_i \leq \tau_j$, with the input parameters fixed throughout:
\[
Q_{\text{cond}}: P(L_{\tau_j} \mid \alpha_{0:\tau_j} \leftarrow c_1, \beta_{0:\tau_j} \leftarrow c_2, \mu_{0:\tau_j} \leftarrow c_3, L_0 \leftarrow c_4, L_{\tau_i} = c_5)
\]

\paragraph{Parameter intervention queries} These queries change the value of an input parameter at time $\tau_i$ and query the queue-length distribution at a later time $\tau_j$. The parameter value after the intervention differs from its initial value at $t = 0$:
\[
Q_{\text{param}}: P(L_{\tau_j} \mid \alpha_{0:\tau_i} \leftarrow c_1, \alpha_{\tau_i:\tau_j} \leftarrow c_1', \beta_{0:\tau_j} \leftarrow c_2, \mu_{0:\tau_j} \leftarrow c_3, L_0 \leftarrow c_4)
\]
where $c_1' \neq c_1$. Analogous queries are constructed for interventions on $\beta$ and $\mu$.

\paragraph{Queue-length intervention queries} These queries set the queue length to a fixed constant at time $\tau_i$ and query the queue-length distribution at time $\tau_j$ ($i < j$):
\[
Q_{\text{qint}}: P(L_{\tau_j} \mid \alpha_{0:\tau_j} \leftarrow c_1, \beta_{0:\tau_j} \leftarrow c_2, \mu_{0:\tau_j} \leftarrow c_3, L_0 \leftarrow c_4, L_{\tau_i} \leftarrow c_5)
\]

\paragraph{Queue-length change intervention queries} These queries modify the queue length at time $\tau_i$ by adding or subtracting a constant, and then query the queue-length distribution at time $\tau_j$ ($i < j$):
\[
Q_{\text{qchg}}: P(L_{\tau_j} \mid \alpha_{0:\tau_j} \leftarrow c_1, \beta_{0:\tau_j} \leftarrow c_2, \mu_{0:\tau_j} \leftarrow c_3, L_0 \leftarrow c_4, L_{\tau_i} \leftarrow L_{\tau_i} \oplus c_5)
\]
where $\oplus \in \{+, -\}$.

\subsection{Evaluation Metrics}
\label{sec:metrics}

Unlike the M/M/1 queue, closed-form expressions for the queue-length distributions under interventions are not available for the G/M/1 queue. We therefore obtain ground-truth distributions by implementing a version of the simulation model that allows the type of interventions listed in the query workload $W$. Next, we run 50{,}000 Monte Carlo replications of the simulation model per query. The resulting empirical distributions serve as the reference against which the MDBN's inferred distributions are compared.

Our primary metric is the Jensen-Shannon distance (JSD), the symmetrized and smoothed square root of the KL divergence, bounded between 0 and 1, with values close to 0 indicating nearly identical distributions. We additionally evaluate the MDBN's accuracy on two summary statistics: the mean $\mathbb{E}[L_{\tau_j}]$ and the inter-quartile range (IQR) of the queue length distribution at query time $\tau_j$, assessed using the mean absolute error (MAE) as the difference between the summary statistic of the MDBN when compared to the estimated ground-truth. This is also visualized via scatter plots of predicted versus ground-truth values. 

\subsection{Gamma/M/1}
\label{sec:gamma-results}

\begin{figure}[htbp]
    \centering
    \begin{subfigure}[b]{0.31\textwidth}
        \centering
        \includegraphics[width=\textwidth]{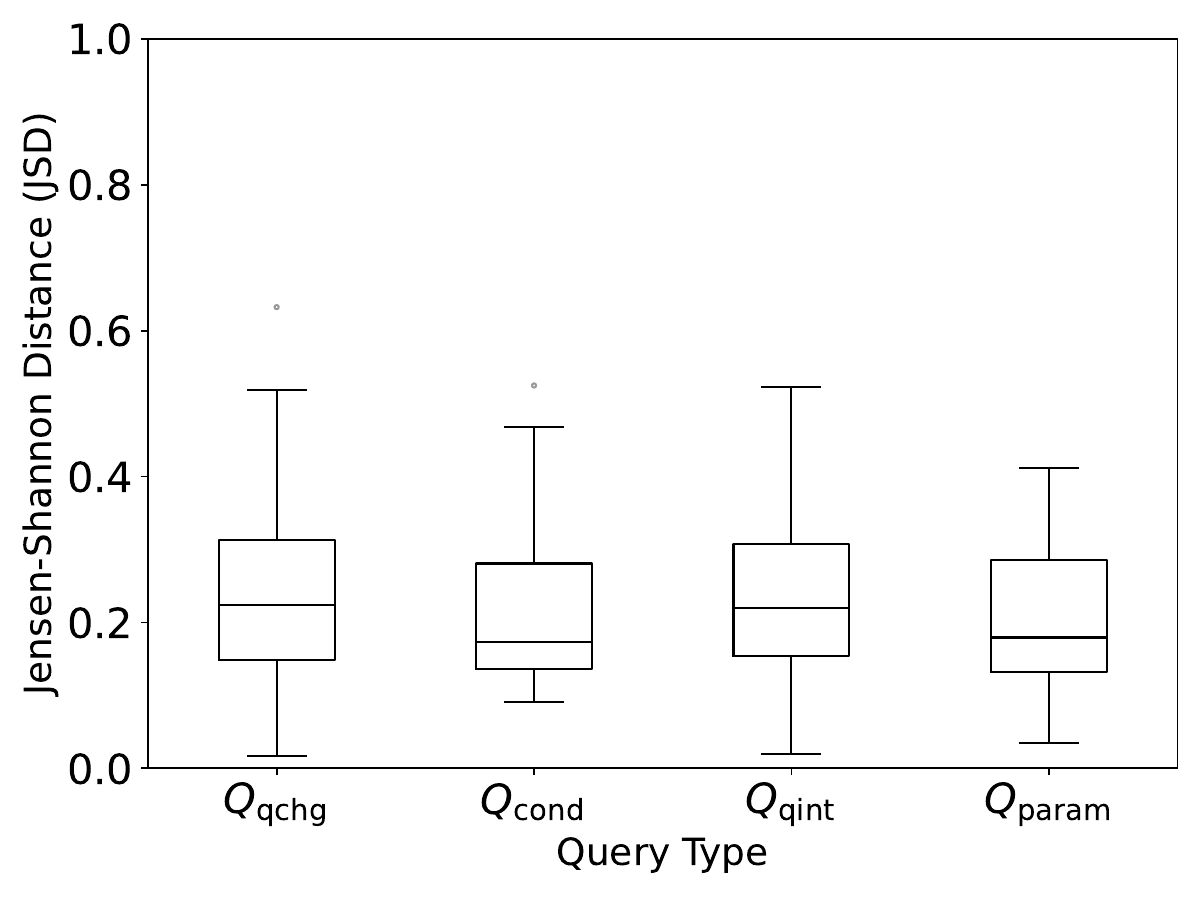}
        \caption{JSD by query type}
        \label{fig:gamma-jsd}
    \end{subfigure}
    \hfill
    \begin{subfigure}[b]{0.31\textwidth}
        \centering
        \includegraphics[width=\textwidth]{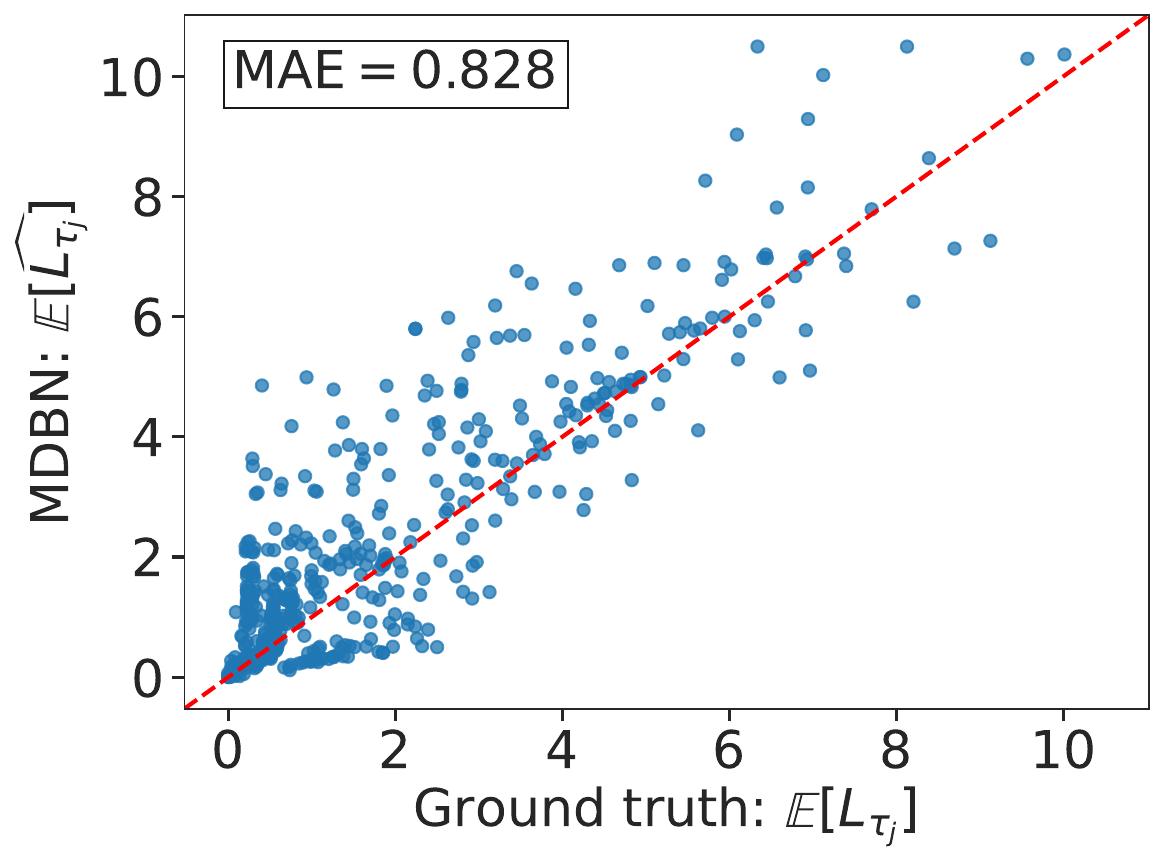}
        \caption{MDBN vs.\ ground-truth $\mathbb{E}[L_{\tau_j}]$}
        \label{fig:gamma-scatter-mean}
    \end{subfigure}
    \hfill
    \begin{subfigure}[b]{0.32\textwidth}
        \centering
        \includegraphics[width=\textwidth]{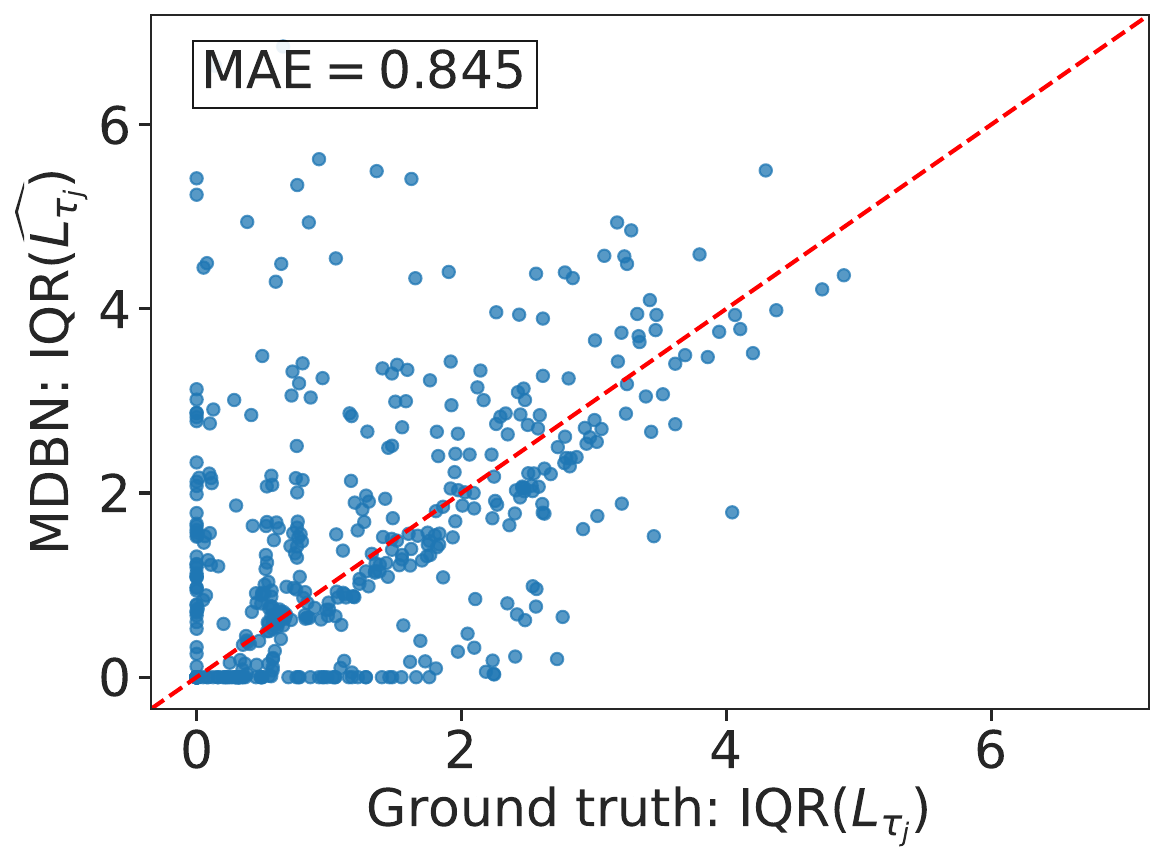}
        \caption{MDBN vs. ground-truth $\mathrm{IQR}(L_{\tau_j})$}
        \label{fig:gamma-scatter-var}
    \end{subfigure}
    \caption{Evaluation of the MDBN on the Gamma/M/1 queue.}
    \label{fig:gamma-results}
\end{figure}

For the Gamma/M/1 distribution, we use the workload $W: \alpha \in \{2.7, 4.3\}, \beta \in \{0.6, 1.2\}, \mu \in \{0.7, 0.9\}$. Figure~\ref{fig:gamma-jsd} plots the JSD across the four query types, with the mean JSD $=0.225$. The scatter plots in Figures~\ref{fig:gamma-scatter-mean} and ~\ref{fig:gamma-scatter-var} show that the predicted means and inter-quartile ranges closely track the ground truth.

\subsection{Weibull/M/1}
\label{sec:weibull-results}

\begin{figure}[htbp]
    \centering
    \begin{subfigure}[b]{0.31\textwidth}
        \centering
        \includegraphics[width=\textwidth]{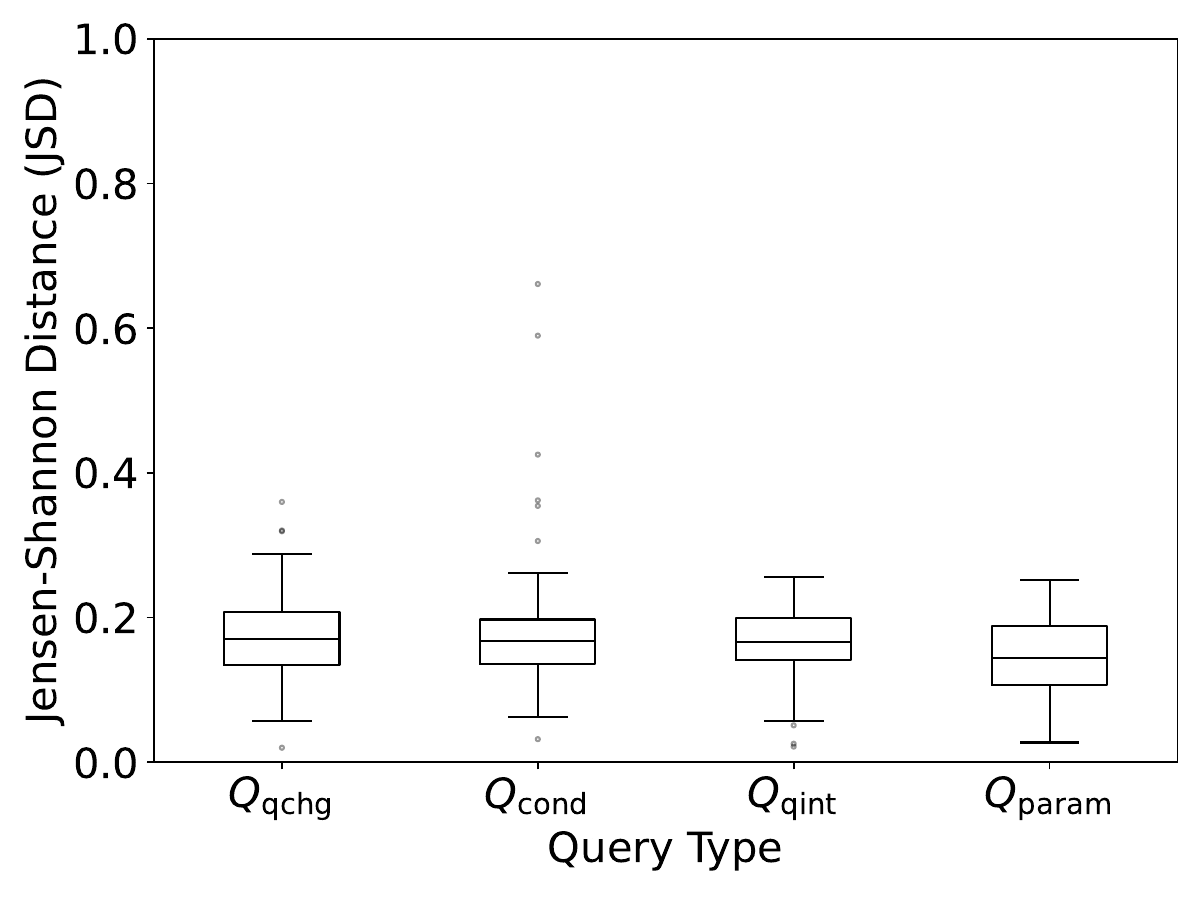}
        \caption{JSD by query type}
        \label{fig:weibull-jsd}
    \end{subfigure}
    \hfill
    \begin{subfigure}[b]{0.31\textwidth}
        \centering
        \includegraphics[width=\textwidth]{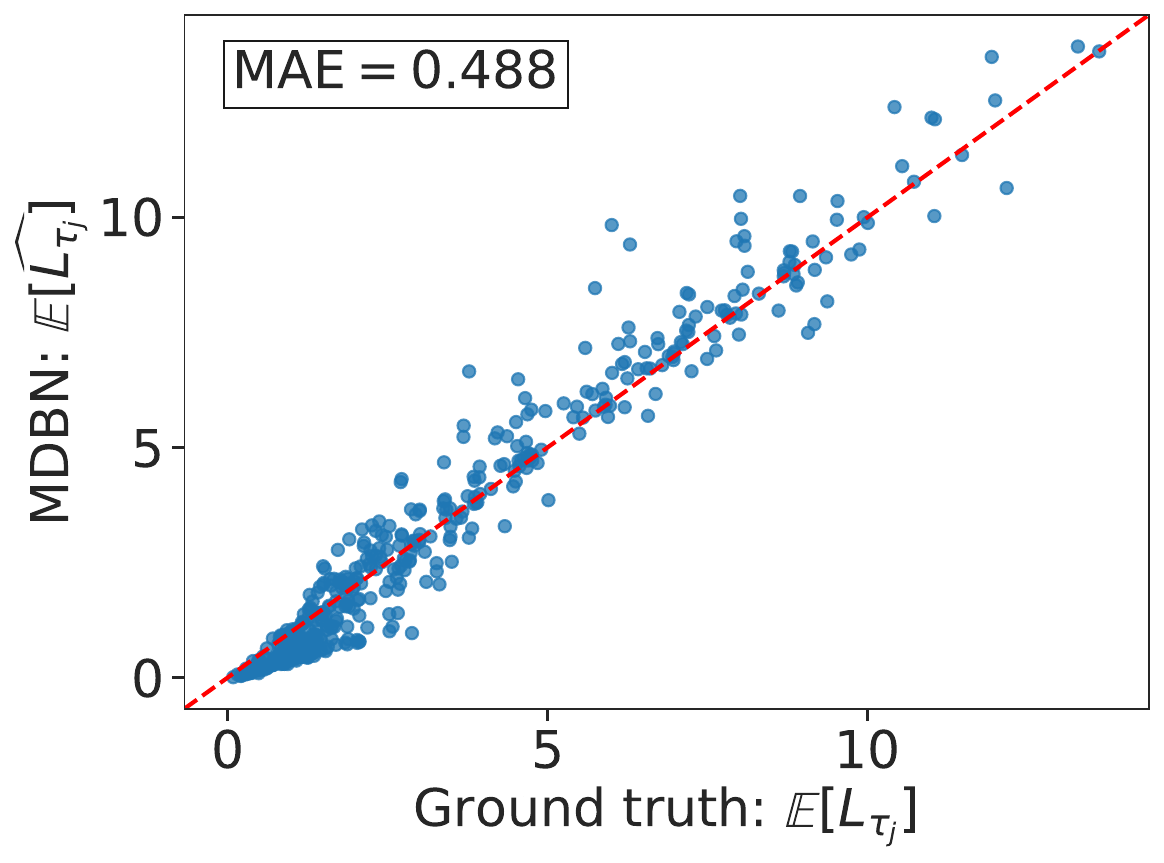}
        \caption{MDBN vs.\ ground-truth $\mathbb{E}[L_{\tau_j}]$}
        \label{fig:weibull-scatter-mean}
    \end{subfigure}
    \hfill
    \begin{subfigure}[b]{0.32\textwidth}
        \centering
        \includegraphics[width=\textwidth]{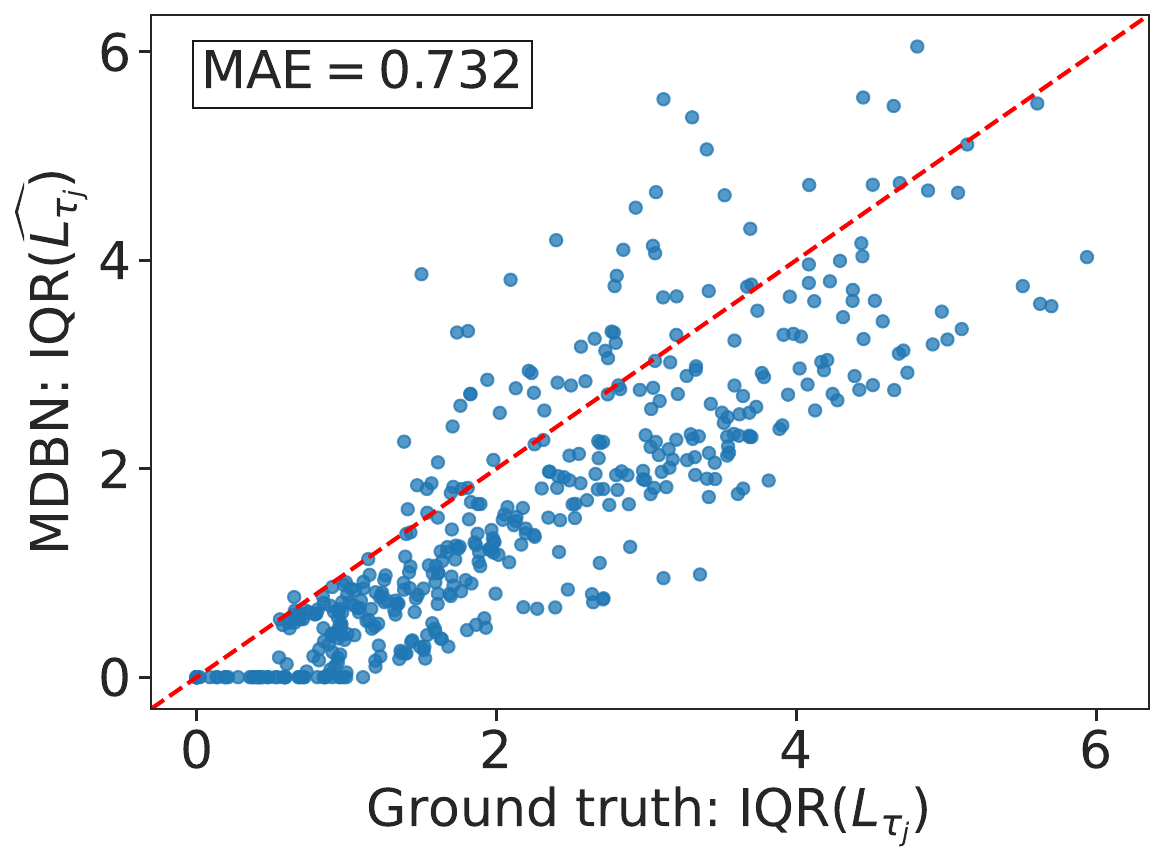}
        \caption{MDBN vs.\ ground-truth $\mathrm{IQR}(L_{\tau_j})$}
        \label{fig:weibull-scatter-var}
    \end{subfigure}
    \caption{Evaluation of the MDBN on the Weibull/M/1 queue.}
    \label{fig:weibull-results}
\end{figure}

For the Weibull/M/1 distribution, we use the workload $W: \alpha \in \{0.9, 1.2\}, \beta \in \{1.5, 2.0\}, \mu \in \{0.8, 1.0\}$. Figure~\ref{fig:weibull-jsd} shows the JSD across query types (mean JSD $=0.166$). The scatter plots in Figures~\ref{fig:weibull-scatter-mean} and~\ref{fig:weibull-scatter-var} show well-calibrated mean (MAE $=0.488$) and IQR predictions (MAE $=0.732$). 

\subsection{Beta/M/1}
\label{sec:beta-experiments}

\begin{figure}[htbp]
    \centering
    \begin{subfigure}[b]{0.31\textwidth}
        \centering
        \includegraphics[width=\textwidth]{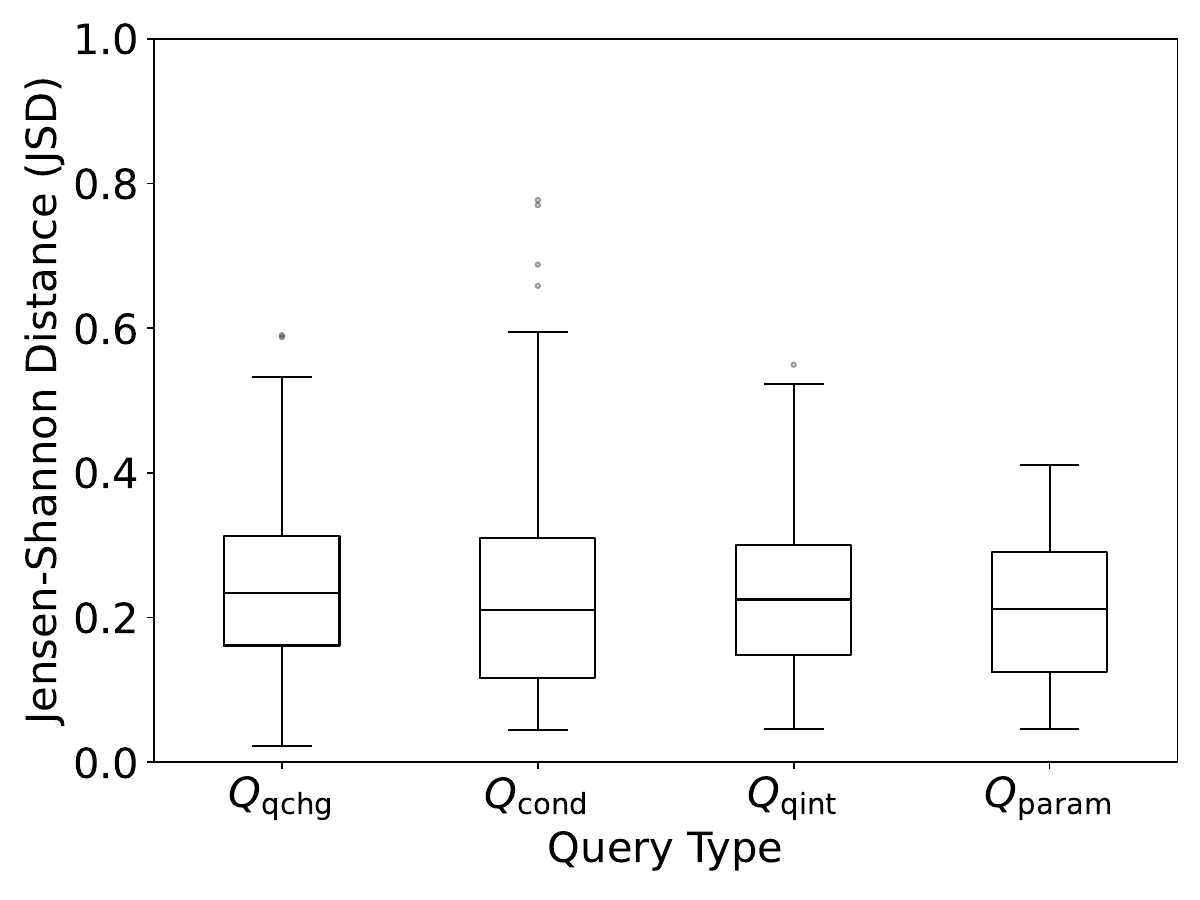}
        \caption{JSD by query type}
        \label{fig:beta-jsd}
    \end{subfigure}
    \hfill
    \begin{subfigure}[b]{0.31\textwidth}
        \centering
        \includegraphics[width=\textwidth]{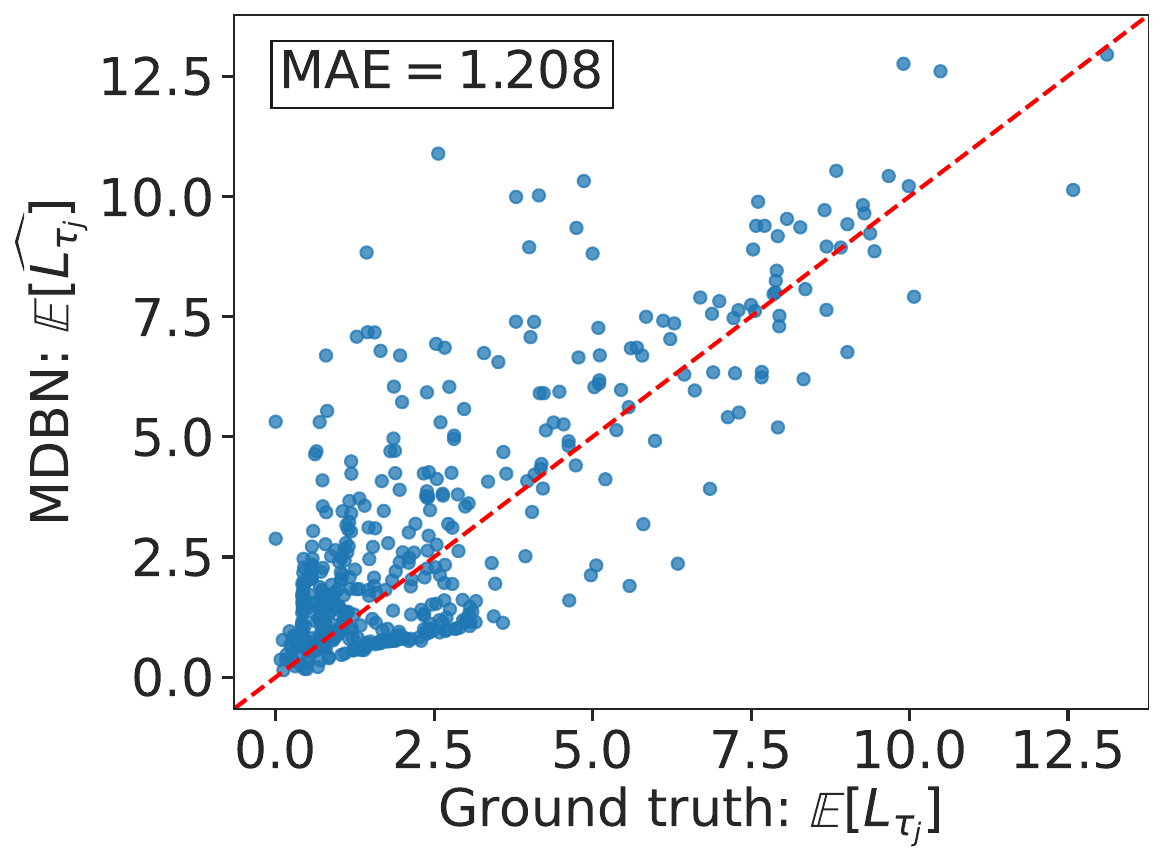}
        \caption{MDBN vs.\ ground-truth $\mathbb{E}[L_{\tau_j}]$}
        \label{fig:beta-scatter-mean}
    \end{subfigure}
    \hfill
    \begin{subfigure}[b]{0.32\textwidth}
        \centering
        \includegraphics[width=\textwidth]{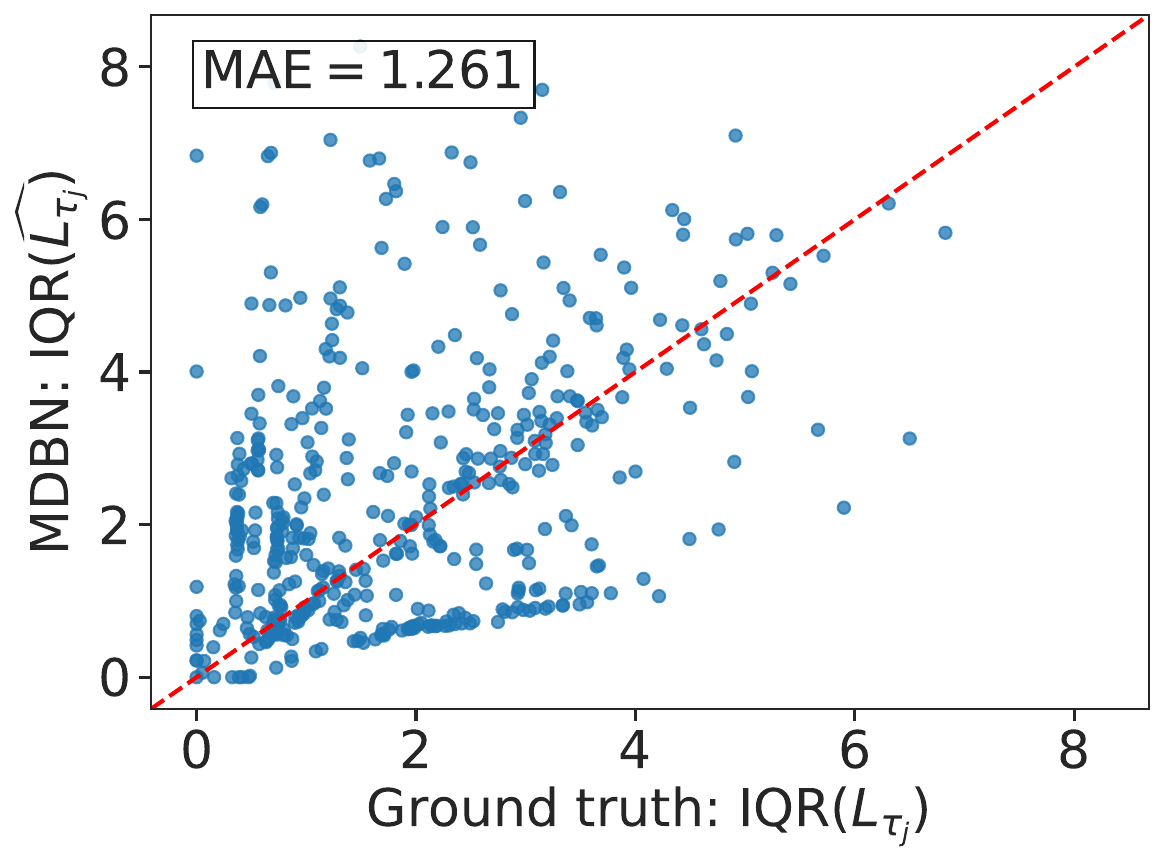}
        \caption{MDBN vs.\ ground-truth $\mathrm{IQR}[L_{\tau_j}]$}
        \label{fig:beta-scatter-var}
    \end{subfigure}
    \caption{Evaluation of the MDBN on the Beta/M/1 queue.}
    \label{fig:beta-results}
\end{figure}

For the Beta/M/1 distribution, we use the workload $W: \alpha \in \{2.5, 1.5\}, \beta \in \{2.5, 4.5\}, \mu \in \{4.5, 5.5\}$. Figure~\ref{fig:beta-jsd} shows the JSD across query types (mean JSD = $0.236$). The Beta distribution presents a more challenging case than the Gamma or Weibull, with higher mean JSD values and greater variability, particularly for interventional queries. The scatter plots in Figure~\ref{fig:beta-scatter-mean} and~\ref{fig:beta-scatter-var} show reasonable agreement for the mean (MAE $=1.208$) and IQR (MAE $=1.261$).

\subsection{Inference Efficiency}
We compare the time to answer PCQs via the MDBN to the time required to obtain ground-truth estimates via Monte Carlo simulation of the (modified) G/M/1 queue, using 50{,}000 sequential replications per query. As shown in Table~\ref{tab:inference-times}, MDBN inference is approximately four orders of magnitude faster than direct simulation across query types and inter-arrival distributions. Even if the 50{,}000 replications were distributed across 1{,}000 parallel processes, the MDBN would still be at least an order of magnitude faster than using the simulation model. Moreover, this comparison does not account for the additional effort required to modify the simulation code to support each intervention type---interventions that the MDBN handles natively.

\begin{table}[htbp]\small
\centering
\caption{Inference time (seconds) for MDBN and ground-truth Monte-Carlo simulations (50,000 replications per query), reported as mean $\pm$ standard deviation across 500 queries.}
\label{tab:inference-times}
\begin{tabular}{l l c c}
\toprule
Inter-Arrival & Query Type & \multicolumn{2}{c}{Inference time (s)}\\ 
Distribution & & MDBN & Ground-truth\\
\midrule
\multirow{4}{*}{Gamma}
 & $Q_{\mathrm{cond}}$   & $0.072 \pm 0.040$ & $540 \pm 299$ \\
 & $Q_{\mathrm{param}}$  & $0.046 \pm 0.031$ & $572 \pm 300$ \\
 & $Q_{\mathrm{qint}}$   & $0.052 \pm 0.028$ & $744 \pm 381$ \\
 & $Q_{\mathrm{qchg}}$   & $0.045 \pm 0.026$ & $606 \pm 306$ \\
\noalign{\smallskip}\hline\noalign{\smallskip}
\multirow{4}{*}{Weibull}
 & $Q_{\mathrm{cond}}$   & $0.065 \pm 0.042$ & $814 \pm 498$ \\
 & $Q_{\mathrm{param}}$  & $0.048 \pm 0.035$ & $910 \pm 448$ \\
 & $Q_{\mathrm{qint}}$   & $0.049 \pm 0.026$ & $788 \pm 327$ \\
 & $Q_{\mathrm{qchg}}$   & $0.039 \pm 0.023$ & $739 \pm 343$ \\
\noalign{\smallskip}\hline\noalign{\smallskip}
\multirow{4}{*}{Beta}
 & $Q_{\mathrm{cond}}$   & $0.048 \pm 0.034$ & $1393 \pm 784$ \\
 & $Q_{\mathrm{param}}$  & $0.032 \pm 0.026$ & $1543 \pm 770$ \\
 & $Q_{\mathrm{qint}}$   & $0.033 \pm 0.017$ & $1656 \pm 855$ \\
 & $Q_{\mathrm{qchg}}$   & $0.030 \pm 0.019$ & $1462 \pm 840$ \\
\bottomrule
\end{tabular}
\end{table}

\section{DISCUSSION}
\label{sec:discussion}
The experimental results demonstrate that MDBNs are capable of producing accurate answers for a range of probabilistic and causal queries on non-Markovian queues, with conditional and parameter intervention queries consistently achieving low JSD across a variety of inter-arrival distributions. The Weibull/M/1 queue yields the best overall performance, while the Beta/M/1 queue proves most challenging, likely due to the bounded support of the Beta distribution requiring a more complex phase-type approximation.

The error observed in the MDBN's inferred distributions arises from two sources: the phase-type approximation of the general inter-arrival distribution, and the MDBN's approximation of the resulting GED-augmented system. Several factors influence the latter. First, the accuracy of the estimated CPD parameters depends on the number of simulation runs $N$ per design point; increasing $N$ may yield more accurate estimates, particularly for states visited infrequently. Second, the choice of sampling interval $\delta$ introduces a tradeoff: while a smaller $\delta$ provides a finer-grained discretization of the underlying CTMC, it also increases the number of time slices over which inference is performed, causing approximation errors to accumulate across successive slices~\cite{doytchinov2010time,boyen2013tractable}. In this paper, we attempt to balance these competing effects by both choosing the optimal sampling interval for a query workload, and by minimizing the number of phases.

\section{CONCLUSION}
\label{sec:conclusion}
We have demonstrated that MDBNs can be extended to non-Markovian queues by adapting the method of phases.
Our experiments on the G/M/1 queue with Gamma, Weibull, and Beta inter-arrival distributions shows that the extended MDBN can produce accurate answers to both conditional and interventional queries, with inference times orders of magnitude faster than Monte Carlo simulation. The techniques introduced here
are not specific to the G/M/1 queue and carry over directly to the G/G/1 setting, where the service-time distribution may also be non-Markovian.

In future work, we plan to focus on scaling the MDBN framework in terms of both learning and inference. In this paper, exact inference via lazy propagation was sufficient, but as the MDBN grows to accommodate more complex systems such as networks of queues, we anticipate that approximate inference techniques will become necessary. We also plan to investigate whether use of the generalized semi-Markov processes framework \cite{shedler1992regenerative} can lead to more compact MDBNs.

\footnotesize

\bibliographystyle{wsc}

\bibliography{references}

\section*{AUTHOR BIOGRAPHIES}

\noindent {\bf \MakeUppercase{Pracheta Amaranath}} is a Ph.D. Candidate at the Manning College of Information and Computer Sciences, University of Massachusetts Amherst, USA. 
Her e-mail address is \email{pboddavarama@umass.edu} and her website is \url{https://prachetaba.github.io}.\\

\noindent {\bf \MakeUppercase{Anant Bhide}} is a Masters student at the Manning College of Information and Computer Sciences, University of Massachusetts Amherst, USA. 
His e-mail address is \email{abhide@umass.edu}.\\

\noindent {\bf \MakeUppercase{David Jensen}} is a Professor at the Manning College of Information and Computer Sciences, University of Massachusetts Amherst, USA.  His email address is \email{jensen@umass.edu}, and his website is \url{https://groups.cs.umass.edu/jensen/}.\\

\noindent {\bf \MakeUppercase{Peter J. Haas}} is a Professor at the Manning College of Information and Computer Sciences, University of Massachusetts Amherst, USA. His e-mail address is \email{phaas@umass.edu} and his website is \url{https://people.cs.umass.edu/~phaas/}.\\

\end{document}

%% file: wscbib.tex
\makeatletter
\let\@internalcite\cite
\def\cite{\def\@citeseppen{-1000}%
    \def\@cite##1##2{(##1\if@tempswa , ##2\fi)}%
    \def\citeauthoryear##1##2##3{##1 ##3}\@internalcite}
\def\citeNP{\def\@citeseppen{-1000}%
    \def\@cite##1##2{##1\if@tempswa , ##2\fi}%
    \def\citeauthoryear##1##2##3{##1 ##3}\@internalcite}
\def\citeN{\def\@citeseppen{-1000}%
    \def\@cite##1##2{##1\if@tempswa, ##2)\else{}\fi}%
    \def\citeauthoryear##1##2##3{##1 (##3)}\@citedata}
\def\citeA{\def\@citeseppen{-1000}%
    \def\@cite##1##2{(##1\if@tempswa , ##2\fi)}%
    \def\citeauthoryear##1##2##3{##1}\@internalcite}
\def\citeANP{\def\@citeseppen{-1000}%
    \def\@cite##1##2{##1\if@tempswa , ##2\fi}%
    \def\citeauthoryear##1##2##3{##1}\@internalcite}
\def\shortcite{\def\@citeseppen{-1000}%
    \def\@cite##1##2{(##1\if@tempswa , ##2\fi)}%
    \def\citeauthoryear##1##2##3{##2 ##3}\@internalcite}
\def\shortciteNP{\def\@citeseppen{-1000}%
    \def\@cite##1##2{##1\if@tempswa , ##2\fi}%
    \def\citeauthoryear##1##2##3{##2 ##3}\@internalcite}
\def\shortciteN{\def\@citeseppen{-1000}%
    \def\@cite##1##2{##1\if@tempswa, ##2\else{}\fi}%
    \def\citeauthoryear##1##2##3{##2 (##3)}\@citedata}
\def\shortciteA{\def\@citeseppen{-1000}%
    \def\@cite##1##2{(##1\if@tempswa , ##2\fi)}%
    \def\citeauthoryear##1##2##3{##2}\@internalcite}
\def\shortciteANP{\def\@citeseppen{-1000}%
    \def\@cite##1##2{##1\if@tempswa , ##2\fi}%
    \def\citeauthoryear##1##2##3{##2}\@internalcite}
\def\citeyear{\def\@citeseppen{-1000}%
    \def\@cite##1##2{(##1\if@tempswa , ##2\fi)}%
    \def\citeauthoryear##1##2##3{##3}\@citedata}
\def\citeyearNP{\def\@citeseppen{-1000}%
    \def\@cite##1##2{##1\if@tempswa , ##2\fi}%
    \def\citeauthoryear##1##2##3{##3}\@citedata}
%
%
%
\def\@citedata{%
    \@ifnextchar [{\@tempswatrue\@citedatax}%
                  {\@tempswafalse\@citedatax[]}%
}

\def\@citedatax[#1]#2{%
\if@filesw\immediate\write\@auxout{\string\citation{#2}}\fi%
  \def\@citea{}\@cite{\@for\@citeb:=#2\do%
    {\@citea\def\@citea{, }\@ifundefined
       {b@\@citeb}{{\bf ?}%
       \@warning{Citation `\@citeb' on page \thepage \space undefined}}%
{\csname b@\@citeb\endcsname}}}{#1}}%

%
\def\@citex[#1]#2{%
\if@filesw\immediate\write\@auxout{\string\citation{#2}}\fi%
  \def\@citea{}\@cite{\@for\@citeb:=#2\do%
    {\@citea\def\@citea{; }\@ifundefined
       {b@\@citeb}{{\bf ?}%
       \@warning{Citation `\@citeb' on page \thepage \space undefined}}%
{\csname b@\@citeb\endcsname}}}{#1}}%

%
\def\@biblabel#1{}
\makeatother



\newdimen\bibindent
\bibindent=0.0em
\def\thebibliography#1{\section*{\refname}\list
   {}{\settowidth\labelwidth{[#1]}
   \leftmargin\parindent
   \itemindent -\parindent
   \listparindent \itemindent
   \itemsep 0pt
   \parsep 0pt}
   \def\newblock{}
   \sloppy
   \sfcode`\.=1000\relax}